\documentclass[10pt,twocolumn,letterpaper]{article}

\usepackage[pagenumbers]{wacv} 

\usepackage{float}
\usepackage[T1]{fontenc}

\definecolor{wacvblue}{rgb}{0.21,0.49,0.74}
\usepackage[pagebackref,breaklinks,colorlinks,allcolors=wacvblue]{hyperref}

\title{SynPanZ: A Synthetic Dataset to Improve Multi-Animal Pose Estimation in Chimpanzees (\textit{Pan troglodytes})}

\author{Flora Talyigas\\
{\tt\footnotesize 2236027@chester.ac.uk}
\and
Lindsay Murray\\
{\tt\footnotesize l.murray@chester.ac.uk}
\and
Sonya Hill\\
{\tt\footnotesize hill.s@chester.ac.uk}
\and
Richard Stocker\\
{\tt\footnotesize r.stocker@chester.ac.uk}
\and
University of Chester, England\\
}

\begin{document}
\maketitle
\begin{abstract}
Pose estimation underpins a wide range of automated monitoring workflows in animal behaviour research. Unfortunately, endangered great apes such as chimpanzees suffer a chronic shortage of the type of hand-annotated training data that has made robust pose estimation in humans and other model organisms possible.
This paper introduces SynPanZ, a synthetically generated dataset for pose estimation in great apes. Our pipeline renders multi-individual scenes from rigged, behaviourally animated chimpanzee models with randomised poses, lighting conditions, environmental contexts, and camera motion, producing animated RGB videos with per-instance 2D keypoint locations without manual labelling. To test whether synthetic augmentation improves cross-dataset generalisation, four DeepLabCut models are trained under matched settings on data sourced from DeepWild with varying real-to-synthetic augmentation ratios. Models are evaluated on ChimpACT, a visually disjoint zoo dataset. We find that augmenting the DeepWild training data with approximately three times as many synthetic frames nearly doubles instance coverage (12.34\% to 22.54\%) while maintaining keypoint localisation accuracy on detected individuals (PCK@0.20: 49.71\% to 52.25\%). Synthetic-only training fails, showing that rendered data complements rather than replaces real annotated data. This work represents a step towards reliable pose estimation for great apes. 

\end{abstract}
    
\section{Introduction}
\label{sec:intro}

Traditionally, animal behaviour datasets are collected by field workers or researchers who observe behaviours in person or record footage for manual annotation. This approach is labour-intensive and inefficient, especially since a single observer is inherently limited to attending to one individual at a time. Manually annotated behavioural datasets are also susceptible to systematic biases 
\cite{kaufman2009can}, and social animals like great apes can alter their behaviour in response to human presence, further distorting the data collected \cite{mcdougall2012passive}.

The automated collection and analysis of behavioural data offers solutions to many of these problems \cite{vogg2025computer}. 
In humans, both RGB-based and skeleton-based methods for action recognition and behavioural analysis are well-established and systematically benchmarked \cite{wang2023comprehensive}.
However, particularly within neuroscience and ethology, pose estimation has emerged as the crucial intermediate step in automated animal behaviour analysis, compressing raw video into skeletal trajectories for downstream analyses \cite{sturman2020deep, hsu2021b, luxem2022identifying, goodwin2024simple}. 

As our closest living relatives, chimpanzees are a uniquely important target for automated behavioural analysis methods \cite{vogg2025computer, ma2026alphachimp}.
Their complex social structures, tool use, and cultural variation make them central to comparative cognition research, yet these same behaviours them make some of the most challenging non-human subjects for pose estimation, and the downstream automated behavioural analyses that pose estimation would facilitate \cite{wiltshire2023deepwild}.

While pose estimation models for humans may be trained on millions of images, publicly available datasets for chimpanzees such as DeepWild \cite{wiltshire2023deepwild} and ChimpACT \cite{ma2023chimpact} contain only 2,000 to 16,000 labelled images. 
In fact, collecting a dataset with coverage comparable to human pose estimation models has been estimated to cost more than \$10~M \cite{bala2020automated}. Issues arising from small existing dataset sizes are further compounded by the fact that chimpanzees inhabit diverse environments ranging from dense tropical forest to open savanna in the wild, and occupy varied indoor and outdoor enclosures in captivity, introducing substantial variation in background context, scene clutter, lighting conditions, camera viewpoint, and number of individuals. As such, building robustly generalising pose estimation models for chimpanzees remains a difficult problem.

A promising strategy for improving pose estimation under these conditions, without incurring the cost of additional manual annotation, is the use of synthetically generated training data. Rendering pipelines allow controlled variation in pose, body appearance, environment, and lighting to be introduced systematically, expanding the effective training distribution at low marginal cost. Prior work on synthetic animal data generation, including the replicAnt pipeline \cite{plum2023replicant}, the PASyn approach \cite{jiang2022pasyn}, and the endangered-species pose estimation guide of Straka et al. \cite{straka2024hitchhikers}, has demonstrated that synthetic data can reduce the hand-annotation burden and improve domain robustness. 

However, to our knowledge, no prior work has applied 3D-rendered synthetic data generation to pose estimation in chimpanzees or other great apes despite the significant lack of hand-labelled data. This means that the question of whether synthetic data offers a meaningful benefit for chimpanzee pose estimation, and specifically whether it supports cross-dataset generalisation across visually diverse environments, remains open.

In this work, we introduce SynPanZ, a new synthetic dataset for training multi-animal chimpanzee pose estimation models. To evaluate the benefit of this dataset, we train several pose estimation models in DeepLabCut \cite{mathis2018deeplabcut} under matched settings, varying the proportion of synthetic and real training data. Real data are sourced from the DeepWild dataset \cite{wiltshire2023deepwild}, consisting of wild chimpanzee footage in forest environments. We evaluate four configurations, a fully real baseline (\textit{1real}), a mixed configuration with approximately the same amount of real and synthetic data (\textit{1real-1syn}), a configuration with approximately three-fold synthetic augmentation (\textit{1real-3syn}), and a fully synthetic configuration (\textit{3syn}). We evaluate all models on the ChimpACT dataset \cite{ma2023chimpact}, which provides a visually distinct test environment, encompassing indoor and outdoor zoo enclosures, man-made structures, vegetation, varying illumination, and both handheld and static camera viewpoints. This cross-dataset protocol directly probes whether synthetic data can improve robustness to novel environments.

Our experimental results show that mixing real and synthetic data improves cross-dataset generalisation, yielding higher average precision and better per-instance coverage compared to the real-only baseline. These benefits increased with synthetic data volume across the configurations tested, even when such generated data was the majority. However, models fine-tuned on synthetic data alone underperform those that utilise much smaller real datasets. These findings suggest that synthetic data is a useful complement to scarce real ape annotations, while also highlighting the need for improved realism and occlusion modelling to achieve robust great ape pose estimation at scale.
\section{Related Work}
\label{sec:prevwork}
\subsection{Pose Estimation in Macaques}
In macaques, the model species among non-human primates, increasing pose estimation efforts have emerged in recent years. Bala and colleagues \cite{bala2020automated} built a markerless 3D motion capture system estimating 3D poses from images depicting freely moving macaques, using 62 cameras around a $2.45\times2.45\times2.75$~m enclosure. They also released the \textbf{OpenMonkeyPose} dataset of 195,228 annotated images. Labuguen and colleagues \cite{labuguen2021macaquepose} released a hand-annotated 2D dataset of macaques in naturalistic scenes across more than 13,000 images, validated using the DeepLabCut algorithm \cite{mathis2018deeplabcut}. To date, this dataset, called \textbf{MacaquePose}, is the standard 2D in-the-wild benchmark. 

More recently, Matsumoto and colleagues \cite{matsumoto2025three} took one of the first steps toward multi-animal video-based pose recognition by developing a multi-view motion capture system to reconstruct 3D poses of monkeys living in groups. Validation showed that it could track and recognise basic social events from multi-view videos. Among the most recent publications, Martini and colleagues published \textbf{BigMaQ} \cite{martini2026bigmaq}, a 3D motion and animation dataset of macaques with avatar representations rather than keypoints only. They released more than 750 scenes of interacting rhesus macaques with recorded skeletal joint rotations and action labels, recorded from 16 calibrated cameras. They report that adding the avatars to video features substantially improves action recognition performance.

\subsection{Available Datasets for Chimpanzees}

Unlike macaques, chimpanzees are an endangered species. 
The majority of behavioural observations of endangered species occur in the wild, where footage from a single handheld camera is the most common form of data collection.

Furthermore, great-ape-specific cognitive and ethical arguments prevent the use of \textit{invasive} techniques (such as adding surface markers to joints) routinely applied to model organisms \cite{eichner2024detailed}.
Even in captivity, multi-view camera setups are often not achievable for various reasons such as associated cost, enclosure limitations, and privacy policies upheld by institutions.

As such, there are currently only two chimpanzee-specific datasets for pose estimation, DeepWild \cite{wiltshire2023deepwild} and ChimpACT \cite{ma2023chimpact}, both of which are hand-annotated 2D multi-animal video datasets captured partially or entirely by handheld cameras. While DeepWild was recorded in the wild, ChimpACT depicts captive animals in both indoor and outdoor zoo enclosures. There are also two notable single-animal image-based primate datasets, OpenMonkeyChallenge \cite{yao2023openmonkeychallenge}, and OpenApePose \cite{desai2023openapepose}. However, these datasets target single-animal pose estimation and consist of cropped images, reducing their practicality for monitoring tasks. This limitation is particularly relevant for video, where motion blur, defocus, and rapid limb movement can further degrade localisation accuracy \cite{luo2017thinslicing, jiao2025optimizing}.

\subsection{Cross-Species Generalisation}
Given this data scarcity, another appealing approach would be cross-species generalisation from pretrained animal or human models. Unfortunately, previous research efforts seem to contradict optimistic expectations on this front. Evidence against straightforward cross-species generalisation comes from Labuguen and colleagues \cite{labuguen2021macaquepose}, who introduced the MacaquePose dataset. Training and testing pose estimation networks on this dataset significantly reduced prediction error (root-mean-square error (RMSE): 7.83 $\to$ 6.02) compared with general multianimal models. 

Another study supporting the need for specialised models is the work of Desai and colleagues \cite{desai2023openapepose} who trained separate HRNet-W48 architectures on human, macaque and ape images, and despite the significant difference in training dataset sizes (human: 262,465; macaque: 56,942; ape: 43,120), ape keypoint accuracy dropped from PCK (percentage of correct keypoints) 87.6\% on the ape-trained model to 56.9\% on the human-trained model and to 58.4\% on the macaque-trained model. 

Yang and colleagues \cite{yang2022apt} have demonstrated that large animal pose estimation and tracking models generalise poorly to monkeys and apes compared to animals with more stereotypical movement patterns and morphological features. One such feature is the dark, low-texture fur found on chimpanzees, leading to visually ambiguous body parts and making correct keypoint-to-individual assignment in group interactions difficult \cite{ma2023chimpact, fuchs2025forest}.

\subsection{Downstream Applications of Pose Estimation}
These difficulties in pose estimation may explain why comparisons of RGB-based vs skeleton-based action recognition methods for chimpanzees show that skeleton-based action recognition models underperform their RGB-based counterparts \cite{ma2023chimpact, fuchs2025forest, ma2026alphachimp}. Indeed, skeleton inputs of poor quality systematically degrade action recognition performance \cite{liu2024enhancing}, making it difficult to attribute the gap to the skeleton-based modality itself. This matters because the human action recognition literature provides strong evidence that, when pose estimation is reliable, skeleton-based representations offer meaningful advantages: they provide a compact, behaviour-centred description of movement that is considerably more robust to subject variation, background shortcuts, illumination change, and viewpoint shift than raw RGB features \cite{wang2023comprehensive, dave2022anubis}. Pose-based representations are therefore a particularly promising direction for cross-dataset behaviour analysis in great apes.
\section{Methods}
\label{sec:methods}

\subsection{Synthetic Data Generation}

\begin{figure*}[htbp]
\centerline{\includegraphics[scale=0.4]{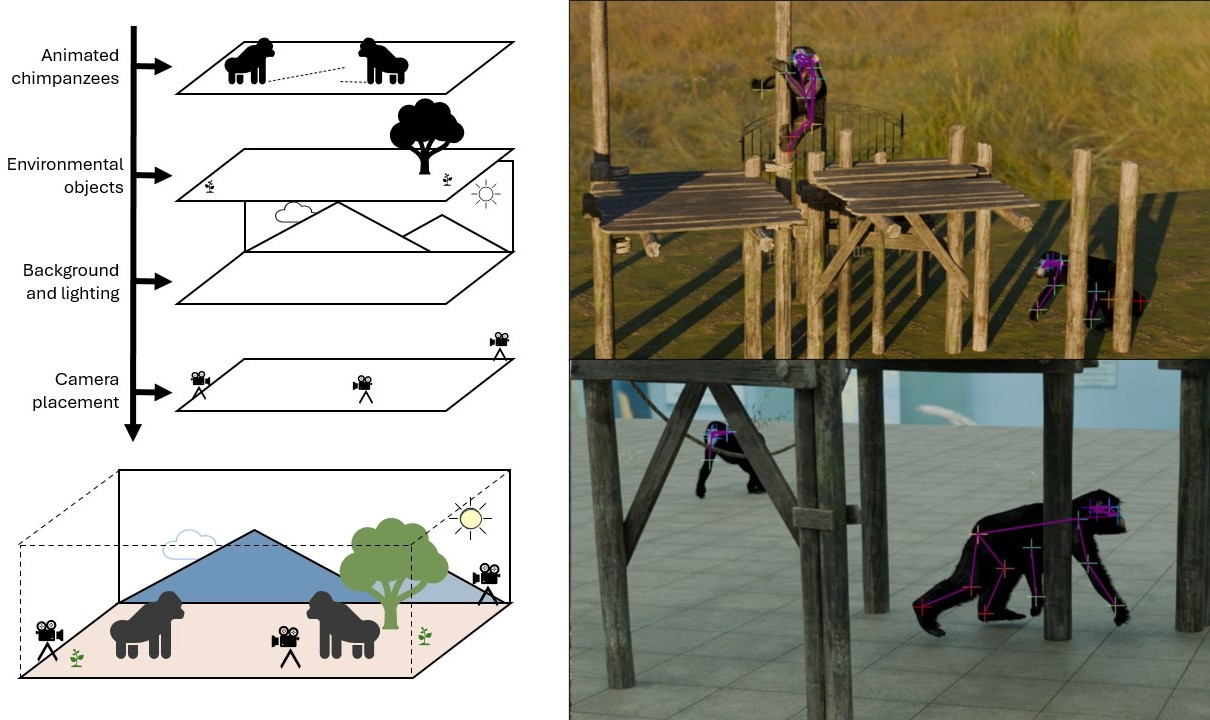}}
\caption{Synthetic generation pipeline diagram (left) and examples of generated images with pose annotations (right).}
\label{fig-datagen}
\end{figure*}

Our primary goal with synthetic data generation is to cover broad visual conditions relevant to chimpanzees, such as indoor rooms, outdoor spaces, climbing structures, vegetation, clutter, changing viewpoint (including both aerial and ground-level footage), and group settings. Our scene-generation pipeline (illustrated in Figure~\ref{fig-datagen}) generates multiple 60-frame clips, exporting frames at $1280 \times 720$ resolution, along with DeepLabCut-compatible keypoint locations and YOLO-style detection labels. The keypoints correspond to the anatomical landmarks used in real-data evaluation, and are exported without visibility flags, as DeepLabCut training does not require them. However, the pipeline optionally exports per-keypoint occlusion states. The current dataset contains 83 clips from 7 scene generations, 5,276 rendered frames in total, and can be scaled without additional manual work. For more details on the SynPanZ dataset, please refer to Table \ref{tab-dataset_stats}.

\begin{table}[t]
  \centering
  \small
  \setlength{\tabcolsep}{4pt}
  \begin{tabular}{ll}
    \toprule
    \multicolumn{2}{l}{\textbf{Scale}} \\
    Generated 3D scenes & 7 \\
    Rendered clips & 83 \\
    Rendered frames & 5,276 \\
    \quad with $\geq$1 visible individual & 5,021 (95.2\%) \\
    Annotated instances & 8,467 \\
    Annotated keypoints & 131,106 \\
    \midrule
    \multicolumn{2}{l}{\textbf{Annotation}} \\
    Keypoints per individual & 18 \\
    Mean keypoints per instance & 15.5 (86.0\%) \\
    Occlusion flags & optional \\
    Label formats & DeepLabCut, YOLO \\
    \midrule
    \multicolumn{2}{l}{\textbf{Scene composition}} \\
    Individuals per scene & 5--10 \\
    Visible individuals per frame & 1.6 (median 2, max 4) \\
    Surface-tone variants & 4 \\
    Behavioural animations & 6 \\
    Environment types & indoor, outdoor \\
    \midrule
    \multicolumn{2}{l}{\textbf{Rendering}} \\
    Resolution & $1280 \times 720$ \\
    Frames per clip & 60 (5 clips at 120) \\
    Camera motion & 50\% tracking / 50\% static \\
    Lighting & sampled HDRI \\
    \bottomrule
  \end{tabular}
  \caption{Statistics of the SynPanZ 1.0 dataset. Instance and
  keypoint counts are computed over individuals falling within the camera frustum. All quantities scale with
  additional pipeline runs, and the tracking/static camera ratio, occlusion export, and per-scene individual count are configurable.}
  \label{tab-dataset_stats}
\end{table}

\begin{figure}
    \centerline{\includegraphics[scale=0.2]{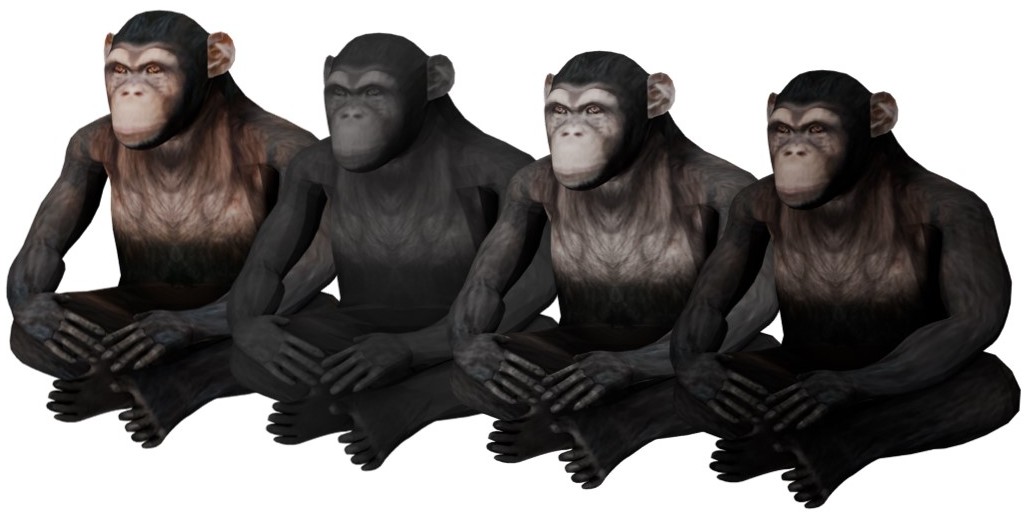}}
    \caption{Four tone variations introduced to the model to increase appearance diversity, captured during the "sitting" animation.}
    \label{fig-tones}
\end{figure}

The pipeline uses a rigged low-poly chimpanzee model built and animated by the authors in Blender \cite{blender}, an open-source 3D modelling and animation package. Four surface-tone variants were introduced to increase appearance variation, as illustrated in Figure~\ref{fig-tones}. Six behavioural animations were implemented: sitting, self-grooming, walking, running, bipedal running, and climbing. Sitting and grooming create compact poses and self-occlusion; walking and running provide quadrupedal motion patterns; bipedal running and climbing add upright configurations.

Scenes are generated in a 20\,m $\times$ 20\,m area with a random number of chimpanzees between 5 and 10. For each individual, the pipeline randomly selects an animation, appearance variant, location, and facing direction, ensuring that chimpanzee meshes do not intersect. Collision-aware object placement then adds indoor or outdoor environmental elements, such as enclosure structures and clutter indoors, and trees, bushes, or rocks outdoors. All environmental objects other than the chimpanzee model and zoo-specific climbing structures were downloaded from Poly Haven's public 3D asset library \cite{polyhaven}.

Lighting and backgrounds are randomly sampled from indoor or outdoor HDRI sets, the ground material is sampled from texture libraries, all sourced from Poly Haven \cite{polyhaven}. To reduce overly sharp synthetic shadows, HDRI lighting is diffused with a monotone grey layer, where the average HDRI brightness defines the layer value. Cameras are randomly placed around the scene and oriented toward a randomly selected chimpanzee. Half track the selected subject through the clip, while the remaining half stay static, creating variation in framing and partial visibility.

\subsection{Network Training}

We emphasise that this experiment is a controlled exploration of the real-to-synthetic training ratio rather than an attempt to establish state-of-the-art chimpanzee pose estimation. All of our configurations share an identical backbone, initialisation, optimiser, schedule, and evaluation protocol, so differences are attributable to training data composition.

\paragraph{Training Procedure}
Four models were trained in DeepLabCut \cite{mathis2018deeplabcut} using a multi-animal ResNet-50 backbone. DeepLabCut is a large open-source toolbox that supports biologists via end-to-end detection, pose estimation, and tracking algorithms. Given its dependable graphical user interface and collection of pretrained models, it is one of the most commonly used pose estimation tools among behaviouralists. Given the widespread familiarity of this platform within our target users, we chose this algorithm for evaluation. Weights were initialised from the SuperAnimal-Quadruped pretrained model \cite{ye2024superanimal}, a large general quadrupedal pose estimation model that supports transfer learning for smaller datasets. Models were trained for 50 epochs using the Adam optimiser with a learning rate of $1\times10^{-5}$, an input crop of $448 \times 448$, and Gaussian plateau heatmaps.

Each model was trained on a desktop workstation equipped with a single NVIDIA RTX 4500 ADA 24\,GB GPU, 16 CPU cores, and 64\,GB of RAM.

\paragraph{Dataset Configurations}
Four training configurations were evaluated. The \textit{1real} dataset only uses DeepWild's annotated frames (1,940 frames), \textit{1real-1syn} adds a roughly equal volume of synthetic data (1,940 real, 1,537 synthetic frames; $\approx$44\% synthetic), \textit{1real-3syn} adds approximately three times as many synthetic frames as real frames (1,940 real, 5,013 synthetic; $\approx$72\% synthetic), and \textit{3syn} is trained on synthetic data only (5,013 frames). The naming convention reflects the approximate data ratios and dataset sizes rather than exact frame counts.

\subsection{Evaluation}
All models were evaluated on the ChimpACT dataset \cite{ma2023chimpact}, specifically on 4,800 labelled frames from 48 videos in diverse environments. For benchmarking, we analysed videos with the DeepWild model as published by the authors. Evaluation was only performed on the 14 keypoints shared between ChimpACT and DeepWild: left/right eye, nose, neck, left/right shoulder, left/right elbow, left/right wrist, left/right knee, and left/right ankle. 

\paragraph{Individual Matching} Predicted and ground-truth individuals are matched per frame via linear-sum assignment on bounding-box IoU. We match on spatial overlap rather than the confidence-ranked greedy matching of the COCO protocol \cite{lin2014coco} because DeepLabCut's confidence outputs are per-keypoint likelihoods rather than per-detection scores. Therefore, the IoU-based assignment component is drawn from the Simple Online and Realtime Tracking (SORT) tracking framework \cite{bewley2016sort}, which introduced per-frame bounding-box IoU matching as an effective strategy for associating detections with tracked instances. Bounding boxes are derived from the predicted keypoints, and following the SORT framework, assignments with IoU < 0.3 are rejected.

\paragraph{Keypoint-level Metrics} Per-keypoint accuracy is measured by two complementary metrics. Normalised prediction error is the Euclidean distance between a predicted keypoint and its ground-truth location, normalised by the square root of the bounding-box area. It serves as the primary metric in the primate pose estimation literature and is our main error measure. 

Keypoint Similarity (KS) maps this scaled Euclidean distance to a [0,1] similarity score via a keypoint-specific falloff constant $\kappa_k$:
\begin{equation}
KS_k = \exp\left(-\frac{d_k^2}{2s^2\kappa_k^2}\right)
\label{eq-ks}
\end{equation}
where $d_k$ is the Euclidean distance between predicted and ground-truth keypoint $k$, $s$ is the ground-truth animal scale defined as the square root of the bounding-box area, and $\kappa_k$ is the keypoint-specific falloff constant. Because chimpanzee-specific redundant annotation statistics are unavailable, human-derived falloff constants are mapped to corresponding anatomical landmarks, as previously done in primates \cite{yao2023openmonkeychallenge}. This mapping is treated as an approximation for comparison to human literature. 

Overall keypoint accuracy is summarised by Percentage of Correct Keypoints (PCK), which is the proportion of predicted keypoints on matched individuals whose normalised error falls below a threshold. Keypoints for which no prediction was returned are excluded from PCK and are instead captured by the coverage metrics. We report PCK@0.05, @0.10, and @0.20 thresholds.

\paragraph{Instance-level Metrics} Instance-level Object Keypoint Similarity (\textit{OKS}) is the sum of KS scores over predicted keypoints, divided by the number of visible ground-truth keypoints for that individual. Using this metric, we compute the mean Average Precision and Recall (\textit{mAP} and \textit{mAR}). Detections are first ranked by mean keypoint confidence, with missing keypoints counted as zero. AP at each OKS threshold uses 101-point interpolated precision, with recall normalised by the total number of ground-truth individuals in the test set, counting undetected individuals as misses. mAP and mAR average over OKS thresholds 0.50--0.95. \textit{Instance coverage} is reported as the fraction of ground-truth individuals matched to a prediction, with \textit{keypoint-level coverage} as a secondary diagnostic.

\paragraph{Regression Analysis} To assess whether prediction accuracy varies with filming conditions, we fit separate ordinary least squares (OLS) regressions for each model configuration, with normalised error as the dependent variable. Five binary video-level predictors are included: location (indoor/outdoor), direct sunlight, background complexity (simple/complex), camera motion (fixed/moving), and the presence of individual clustering, each coded manually. Videos with no matched detections were dropped.
\section{Results}

\subsection{Keypoint Localisation Accuracy}

\begin{figure*}[htbp]
\centerline{\includegraphics[scale=0.6]{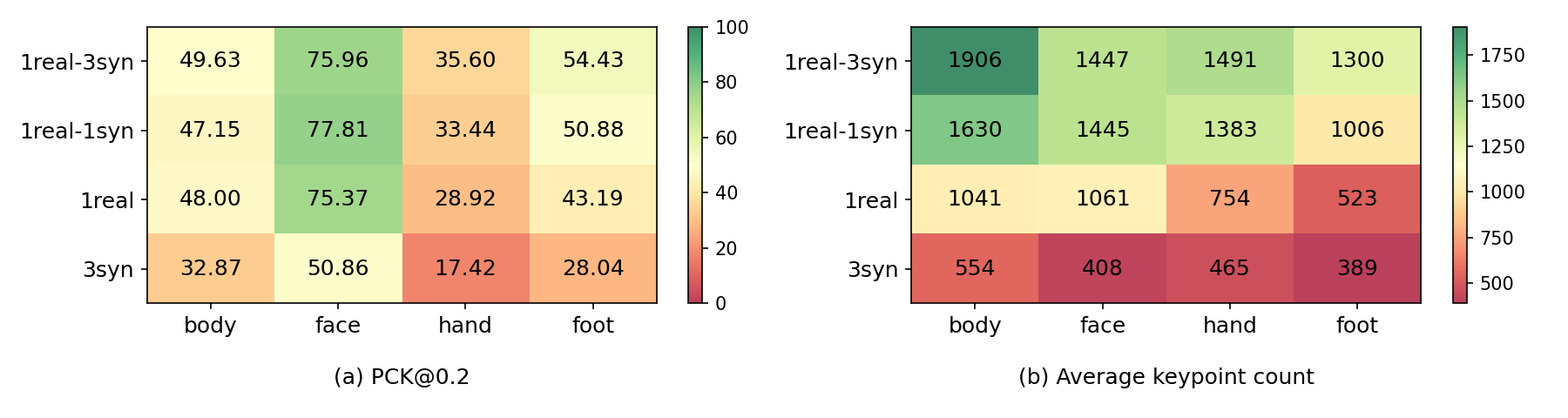}}
\caption{PCK@0.20 (\%) accuracy (a) and mean predicted keypoint count per group (b) across all four model configurations.}
\label{fig-heatmaps}
\end{figure*}

\begin{table*}[htbp]
    \begin{center}
        \begin{tabular}{|l|ccc|cccc|}
            \hline
            Model        & PCK@0.20        & PCK@0.10       & PCK@0.05     & mAP & AP$_{.50}$ & AP$_{.75}$ & mAR \\
            \hline
            1real        & 49.71           & 30.85           & \textbf{17.46}   & 0.001 & 0.010 & 0.000 & 0.002 \\
            1real-1syn & 50.99           & 30.10           & 15.71              & 0.009 & 0.058 & 0.000 & 0.033 \\
            1real-3syn & \textbf{52.25}  & \textbf{31.73}  & 16.07          & \textbf{0.040} & \textbf{0.248} & \textbf{0.005} & \textbf{0.077} \\
            3syn       & 30.61           & 13.93           & 5.50              & 0.000 & 0.000 & 0.000 & 0.000 \\
            \hline
            DeepWild    & 71.29            & 46.17           & 27.15             & 0.013 & 0.124 & 0.000 & 0.003 \\
            \hline
        \end{tabular}
        \caption{Summary of keypoint localisation performance across all model configurations. All metrics are shown in percentages (\%)}
        \label{tab-performance}
    \end{center}
\end{table*}

Table \ref{tab-performance} reports localisation accuracy at three normalised thresholds for all four training configurations, along with the DeepWild model as an outside baseline. 

Both mixed-data configurations outperform the \textit{1real} baseline at PCK@0.20 (50.99\% and 52.25\%, respectively, vs. 49.71\%). At PCK@0.10, however, only \textit{1real-3syn} outperforms the baseline (31.73\% vs. 30.85\%). This improvement with increasing synthetic volume (\textit{1real} $\rightarrow$ \textit{1real-1syn} $\rightarrow$ \textit{1real-3syn}) is consistent with, though not established by, favourable scalability of synthetic augmentation. The \textit{3syn} configuration, trained exclusively on synthetic data, performs substantially worse than all mixed-data models across all thresholds (PCK@0.20: 30.61\%; PCK@0.10: 13.93\%; PCK@0.05: 5.50\%). None of our models reached the accuracy of the DeepWild model (PCK@0.20: 71.29\%; PCK@0.10: 46.17\%; PCK@0.05: 27.15\%).

Per-group accuracy, broken down into \textit{face}, \textit{body}, \textit{hand}, and \textit{foot} keypoints, is visualised in Figure~\ref{fig-heatmaps}. The \textit{1real-3syn} model achieves the highest PCK@0.20 in the body, hand, and foot anatomical groups, with gains observed across all body parts relative to the \textit{1real} baseline. For face-related keypoints, the \textit{1real-1syn} model reaches the highest PCK@0.20 score. The largest absolute improvement is recorded for foot keypoints (43.19\% $\to$ 54.43\%), followed by hand keypoints (28.92\% $\to$ 35.60\%). 

\subsection{Coverage}
Beyond per-keypoint accuracy, a critical practical consideration is instance coverage, defined as the proportion of individuals in the video for which any prediction is produced, reported in Table~\ref{tab-coverage}. The \textit{1real-3syn} model raises instance coverage from 12.34\% to 22.54\% relative to the \textit{1real} baseline, nearly doubling the number of detected individuals. On the other hand, using synthetic data exclusively (\textit{3syn}) recovers the fewest individuals from the trained models, with 6.96\% instance coverage. DeepWild recovers the fewest individuals overall, with 3.16\% of the ground-truth individuals receiving a prediction.

Because PCK and normalised keypoint error are calculated only for matched instances, the models are evaluated on different subsets of the test set. The \textit{1real-3syn} model is evaluated on over twice as many individuals as the \textit{1real} baseline, including individuals that the baseline fails to detect. The accuracy comparison therefore provides a conservative test of our claim, as the augmented model achieves higher PCK@0.20 despite being evaluated on a larger and presumably more challenging set of detected individuals. The same asymmetry applies in the opposite direction to DeepWild, whose accuracy and keypoints-per-instance are computed over 3.16\% of ground-truth individuals.

Keypoint-per-instance coverage represents the average portion of predicted keypoints for a predicted individual. Here, the DeepWild model holds the highest score, as each predicted instance retains 66.05\% of its keypoints. Our models follow closely, with \textit{1real-1syn} at 64.56\%, \textit{1real-3syn} at 63.48\%, and \textit{1real} at 62.58\%. Again, \textit{3syn} underperforms compared to the other models, with an average of 60.59\% of keypoints yielding predictions for an individual.

Examining keypoint detection counts grouped by body parts (Figure~\ref{fig-heatmaps} (b)) further reveals that synthetic data augmentation does not increase keypoint distribution equally. From the \textit{1real} baseline to \textit{1real-3syn}, the total coverage increases by +85\% (7.72\% $\to$ 14.31\%, Table~\ref{tab-coverage}). However, foot keypoint coverage increased by 148\%, while face keypoints grew by only 36\%. After synthetic augmentation, predictions are allocated more evenly across groups, with the biggest improvements in keypoint presence seen in the areas where coverage was originally the worst.

\begin{table}[htbp]
    \begin{center}
        \begin{tabular}{|l|ccc|}
            \hline
            Model       & Instance & Kp-per-instance & Total \\
            \hline
            1real       & 12.34             & 62.58     &   7.72 \\
            1real-1syn  & 19.65             & \textbf{64.56} & 12.69 \\
            1real-3syn  & \textbf{22.54}   & 63.48 & \textbf{14.31} \\
            3syn       & 6.96              & 60.59 & 4.22 \\
            \hline
            DeepWild    & 3.16              & 66.05 & 2.09 \\
            \hline
        \end{tabular}
        \caption{Prediction coverage across models. \textit{Instance coverage} is the portion of individuals with any predictions returned, \textit{kp-per-instance coverage} is the average proportion of predicted keypoints for an accepted instance, while \textit{total coverage} is the percentage of all ground-truth keypoints with valid predictions.}
        \label{tab-coverage}
    \end{center}
\end{table}

\subsection{Overall Performance}
Although mAP is not typically reported for primate pose estimation, we include it to enable comparison with the human literature. Because recall is normalised by all ground-truth individuals, mAP in this setting is dominated by coverage. 1real-3syn achieves the highest mAP (0.040) and mAR (0.077), with mAP values of 1real (0.001) and 3syn (0.000).

\begin{table*}[htbp]
    \begin{center}
        \begin{tabular}{|l|ccccc|cc|}
            \hline
            Model & Outdoor & Sunlight & Simple bg. & Cam.\ motion & Clustering & $R^2$ & $p$ \\
            \hline
            1real        & $-$0.018 & $+$0.170$^{***}$ & $-$0.023 & $-$0.015 & $+$0.088$^{**}$ & 0.468 & $<$0.001 \\
            1real-1syn & $-$0.006 & $+$0.072          & $-$0.017 & $-$0.003 & $+$0.089$^{*}$  & 0.116 & 0.386 \\
            1real-3syn & $-$0.048 & $+$0.124          & $-$0.010 & $-$0.011 & $+$0.133$^{**}$ & 0.224 & 0.057 \\
            3syn       & $-$0.053 & $+$0.171          & $+$0.029 & $-$0.020 & $+$0.157$^{*}$  & 0.207 & 0.183 \\
            \hline
            \multicolumn{8}{c}{}\\
            \multicolumn{8}{c}{Significance: $^{*}p<0.05$, $^{**}p<0.01$, $^{***}p<0.001$.} \\
        \end{tabular}
        \caption{Multiple regression coefficients ($\beta$) for environmental predictors of normalised mean error, per model configuration.}
        \label{tab-ols}
    \end{center}
\end{table*}

\subsection{Effect of Environmental Conditions}
To investigate whether prediction error varies systematically with filming conditions, we regressed the scaled prediction error against five video-level environmental variables: location, presence of direct sunlight, background complexity, camera motion, and individual clustering.

The results are in Table~\ref{tab-ols}. For the \textit{1real} model, the regression explains a substantial proportion of error variance ($R^2 = 0.468$, $F(5,39) = 6.87$, $p < 0.001$), with two predictors reaching significance. Direct sunlight is associated with a marked increase in normalised error ($\beta = 0.170$, $p<0.001$). This association may reflect reduced keypoint visibility under high-contrast lighting. Individual clustering also significantly increases error ($\beta = 0.088$, $p < 0.001$), consistent with the well-documented challenge of multi-instance occlusion in pose estimation. No other variable reached significance.

For the mixed-data configurations, the overall model fits deteriorate considerably. The \textit{1real-1syn} and \textit{3syn} regressions are statistically not significant ($p = 0.386$; $p = 0.183$, respectively), and the \textit{1real-3syn} model only approaches significance ($F(5,41) = 2.36$, $p = 0.057$, $R^2 = 0.224$). The drop in $R^2$ from 0.468 (\textit{1real}) to 0.224 (\textit{1real-3syn}) indicates that synthetic augmentation reduces the degree to which filming conditions explain prediction error. Clustering remains a significant or near-significant predictor.

\section{Discussion}

This work introduces SynPanZ, a synthetic dataset for multi-animal pose estimation in chimpanzees, and investigates whether the augmentation of real data with synthetically generated data can improve chimpanzee pose estimation robustness in multi-instance videos, a domain where labelled real-world data is scarce and costly to produce. Experiments across four training configurations (real-only, two mixed, and synthetic-only datasets) yielded several findings with practical and methodological significance.

\paragraph{An Increase in Coverage}  We find that the primary contribution of synthetic augmentation is the increase in instance coverage. The \textit{1real-3syn} model nearly doubles instance coverage compared with the real-only baseline (12.34\% to 22.54\%) and achieves a seven-fold increase against DeepWild (3.16\% to 22.54\%). Accuracy on detected instances remains broadly comparable across experimental configurations, although each model underperforms relative to DeepWild (PCK@0.20: 71.29\%). Taken together, performance improves from \textit{1real} to \textit{1real-3syn}, as coverage nearly doubles (12.34\% $\to$ 22.54\%) while PCK@0.20 on detected instances rises slightly (49.71\% $\to$ 52.25\%) and keypoints-per-instance holds steady (62.58\% $\to$ 63.48\%). We see only a marginal decrease at the strictest threshold (PCK@0.05), consistent with the known behaviour of synthetic training data in fine-grained localisation tasks. Low-poly meshes capture body geometry sufficiently for coarse placement, but the absence of realistic surface detail means the model may learn to place keypoints relative to geometric rather than textural cues, introducing small systematic offsets \cite{zuffi2019sydog}.

\paragraph{Synthetic Only?}
The underperformance observed in the synthetic-only model may reflect the synthetic-to-real domain gap in the absence of any real-data anchor. Without exposure to individual variation, natural lighting, and video compression artefacts, the model learns pose representations specific to the synthetic dataset. During inference on real footage, these representations fail to activate reliably, producing low detection rates and collapsing mAP.

\paragraph{Environmental Factors}
We also found reduced explanatory power of environmental factors as synthetic data is added ($R^2$ dropping from 0.468 to 0.224), suggesting that synthetic augmentation shifts error from being environmentally predictable to being more idiosyncratic. Nevertheless, clustering consistently predicts increased error across all model configurations, suggesting that occlusion from conspecifics is a fundamental challenge that synthetic augmentation has not yet resolved. 

Future work should therefore prioritise generating synthetic scenes with higher degrees of inter-individual occlusion, such as stereotypical social interactions in the species (e.g. food sharing, infant carrying, and grooming).

\paragraph{Comparison to Existing Solutions}
The DeepWild dataset and model \cite{wiltshire2023deepwild} represent an interesting point of comparison to our pose estimation models. Although all our models are built on the same 'real' dataset as the DeepWild model, that system's accuracy is notably higher than ours, while also retaining more keypoints per instance. 

However, detected individuals are sparse, at 3.16\% instance coverage. This could be due to a longer training period, setting stronger expectations of a forest environment than in our \textit{1real} model, or perhaps transfer learning has been initiated from a different model. Regardless of the cause, the difference in instance coverage strongly influences the instance-level comparison (mAP: 0.013). Since recall is normalised by every ground-truth individual, a model that detects 3.16\% of individuals, and 66\% of keypoints at each instance, cannot exceed an AP of 2.09\% at any OKS threshold, however accurate its detections. The higher mAP of \textit{1real-3syn} (0.040) should therefore be read primarily as a consequence of its higher coverage, as on the individuals it does detect, DeepWild remains the more accurate model. The low absolute mAP relative to the human literature reflects this coverage ceiling.

\paragraph{Limitations and Future Work}
On the evaluation side, each configuration was trained with a single seed. Coverage and accuracy are reported without a confidence threshold. A threshold sweep would be required to separate improvements in localisation from shifts in confidence calibration. Finally, all models were initialised from SuperAnimal-Quadruped and fine-tuned briefly at a low learning rate, a conservative transfer-learning regime that plausibly contributes to the low performance of our configurations. Future work will prioritise longer training regimens and try various architectures to gain further insight into the limiting factors of synthetic data augmentation.

In terms of the dataset itself, the current low-poly chimpanzee mesh captures approximate body geometry but cannot reproduce fine-grained visual detail such as fur texture, skin variation, or anatomical individuality, nor does the renderer fully capture environmental variation such as zoo-glass reflections, dense vegetation, motion blur, or the irregular dynamics of handheld wildlife footage. Future work should pursue more anatomically detailed meshes, a richer behavioural animation set, improved physical contact between animals, and more realistic environments. New versions of the SynPanZ dataset will bring improvements along these axes, which are expected to narrow the domain gap for better transfer to downstream applications.

{
    \small
    \bibliographystyle{ieeenat_fullname}
    \bibliography{main}

@String(IJCV = {Int. J. Comput. Vis.})

@String(CVPR= {IEEE Conf. Comput. Vis. Pattern Recog.})

@String(ECCV= {Eur. Conf. Comput. Vis.})

@String(BMVC= {Brit. Mach. Vis. Conf.})

@String(ICIP = {IEEE Int. Conf. Image Process.})

@String(AAAI = {AAAI})

@String(IJCV  = {IJCV})

@String(CVPR  = {CVPR})

@String(ECCV  = {ECCV})

@String(BMVC  =	{BMVC})

@String(ICIP  = {ICIP})

@InProceedings{lin2014coco,
author={Lin, Tsung-Yi
and Maire, Michael
and Belongie, Serge
and Hays, James
and Perona, Pietro
and Ramanan, Deva
and Doll{\'a}r, Piotr
and Zitnick, C. Lawrence},
editor={Fleet, David
and Pajdla, Tomas
and Schiele, Bernt
and Tuytelaars, Tinne},
title={Microsoft {COCO}: Common Objects in Context},
booktitle={Computer Vision -- ECCV 2014},
year={2014},
publisher={Springer International Publishing},
address={Cham},
pages={740--755},
isbn={978-3-319-10602-1},
doi = {https://doi.org/10.1007/978-3-319-10602-1_48}
}

@article{mathis2018deeplabcut,
  title={{DeepLabCut}: markerless pose estimation of user-defined body parts with deep learning},
  author={Mathis, Alexander and Mamidanna, Pranav and Cury, Kevin M. and Abe, Taiga and Murthy, Venkatesh N. and Mathis, Mackenzie Weygandt and Bethge, Matthias},
  journal={Nature Neuroscience},
  volume={21},
  pages={1281--1289},
  year={2018},
  doi={https://doi.org/10.1038/s41593-018-0209-y}
}

@article{wiltshire2023deepwild,
author = {Wiltshire, Charlotte and Lewis-Cheetham, James and Komedov{\'a}, Viola and Matsuzawa, Tetsuro and Graham, Kirsty E and Hobaiter, Catherine},
title = {{DeepWild}: Application of the pose estimation tool DeepLabCut for behaviour tracking in wild chimpanzees and bonobos},
journal = {Journal of Animal Ecology},
volume = {92},
number = {8},
pages = {1560--1574},
doi = {https://doi.org/10.1111/1365-2656.13932},
year = {2023}
}

@article {desai2023openapepose,
title = {{OpenApePose}, a database of annotated ape photographs for pose estimation},
author = {Desai, Nisarg and Bala, Praneet and Richardson, Rebecca and Raper, Jessica and Zimmermann, Jan and Hayden, Benjamin},
editor = {Kalan, Ammie K and Perry, George H},
volume = {12},
year = {2023},
pages = {RP86873},
doi = {https://doi.org/10.7554/eLife.86873.3},
journal = {eLife},
issn = {2050-084X},
publisher = {eLife Sciences Publications, Ltd},
}

@inproceedings{ma2023chimpact,
 author = {Ma, Xiaoxuan and Kaufhold, Stephan and Su, Jiajun and Zhu, Wentao and Terwilliger, Jack and Meza, Andres and Zhu, Yixin and Rossano, Federico and Wang, Yizhou},
 booktitle = {Advances in Neural Information Processing Systems},
 doi = {https://doi.org/10.52202/075280-1196},
 editor = {A. Oh and T. Naumann and A. Globerson and K. Saenko and M. Hardt and S. Levine},
 pages = {27501--27531},
 publisher = {Curran Associates, Inc.},
 title = {{ChimpACT}: A Longitudinal Dataset for Understanding Chimpanzee Behaviors},
 url = {https://proceedings.neurips.cc/paper_files/paper/2023/file/57a95cd3898bf4912269848a01f53620-Paper-Datasets_and_Benchmarks.pdf},
 volume = {36},
 year = {2023}
}

@article{plum2023replicant,
  title={{replicAnt}: a pipeline for generating annotated images of animals in complex environments using Unreal Engine},
  author={Plum, Fabian and Bulla, Ren{\'e} and Beck, Hendrik K and Imirzian, Natalie and Labonte, David},
  journal={Nature Communications},
  volume={14},
  number={7195},
  year={2023},
  doi={https://doi.org/10.1038/s41467-023-42898-9}
}

@inproceedings{jiang2022pasyn,
  title={Prior-Aware Synthetic Data to the Rescue: Animal Pose Estimation with Very Limited Real Data},
  author={Jiang, Le and Liu, Shuangjun and Bai, Xiangyu  and Ostadabbas, Sarah},
  booktitle={The British Machine Vision Conference (BMVC)},
  month={11},
  year={2022},
  url={https://arxiv.org/abs/2208.13944}
}

@INPROCEEDINGS{straka2024hitchhikers,
  author={Straka, Jakub and Hruz, Marek and Picek, Lukas},
  booktitle={2024 IEEE/CVF Winter Conference on Applications of Computer Vision Workshops (WACVW)}, 
  title={The Hitchhiker's Guide to Endangered Species Pose Estimation}, 
  year={2024},
  pages={41--50},
  doi={https://doi.org/10.1109/WACVW60836.2024.00012}}

@article{yao2023openmonkeychallenge,
  title={{OpenMonkeyChallenge}: Dataset and Benchmark Challenges for Pose Estimation of Non-human Primates},
  author={Yao, Yuan and Bala, Praneet and Mohan, Abhiraj and Bliss-Moreau, Eliza and Coleman, Kristine and Freeman, Sienna M and Machado, Christopher J and Raper, Jessica and Zimmermann, Jan and Hayden, Benjamin Y and Park, Hyun Soo},
  journal={International Journal of Computer Vision},
  volume={131},
  pages={243--258},
  year={2023},
  doi={https://doi.org/10.1007/s11263-022-01698-2}
}

@ARTICLE{labuguen2021macaquepose,
AUTHOR={Labuguen, Rollyn  and Matsumoto, Jumpei  and Negrete, Salvador Blanco  and Nishimaru, Hiroshi  and Nishijo, Hisao  and Takada, Masahiko  and Go, Yasuhiro  and Inoue, Ken-ichi  and Shibata, Tomohiro },  
TITLE={{MacaquePose}: A Novel “In the Wild” Macaque Monkey Pose Dataset for Markerless Motion Capture},
JOURNAL={Frontiers in Behavioral Neuroscience},
volume={14},
pages={581154},
YEAR={2021},
DOI={https://doi.org/10.3389/fnbeh.2020.581154},
ISSN={1662-5153},
}

@INPROCEEDINGS{luo2017thinslicing,
  author={Song, Jie and Wang, Limin and Van Gool, Luc and Hilliges, Otmar},
  booktitle={2017 IEEE Conference on Computer Vision and Pattern Recognition (CVPR)}, 
  title={Thin-Slicing Network: A Deep Structured Model for Pose Estimation in Videos}, 
  year={2017},
  pages={5563--5572},
  doi={10.1109/CVPR.2017.590}}

@article{dave2022anubis,
title = {Representation-centric survey of supervised skeletal action recognition and the new benchmark},
journal = {Pattern Recognition},
pages = {114140},
year = {2026},
issn = {0031-3203},
doi = {https://doi.org/10.1016/j.patcog.2026.114140},
author = {Yang Liu and Jiyao Yang and Madhawa Perera and Pan Ji and Dongwoo Kim and Min Xu and Tianyang Wang and Saeed Anwar and Tom Gedeon and Lei Wang and Zhenyue Qin}
}

@ARTICLE{wang2023comprehensive,
  author={Wang, Cailing and Yan, Jingjing},
  journal={IEEE Access}, 
  title={A Comprehensive Survey of {RGB}-Based and Skeleton-Based Human Action Recognition}, 
  year={2023},
  volume={11},
  pages={53880-53898},
  doi={10.1109/ACCESS.2023.3282311}
}

@article{fuchs2025forest,
  title={From Forest to Zoo: Great Ape Behavior Recognition with {ChimpBehave}},
  author={Fuchs, Michael and Genty, Emilie and Bangerter, Adrian and Zuberb{\"u}hler, Klaus and Odobez, Jean-Marc and Cotofrei, Paul},
  journal={International Journal of Computer Vision},
  volume={133},
  number={10},
  pages={6668--6688},
  year={2025},
  publisher={Springer}
}

@article{ma2026alphachimp,
  title={{AlphaChimp}: Tracking and Behavior Recognition of Chimpanzees},
  author={Ma, Xiaoxuan and Lin, Yutang and Xu, Yuan and Kaufhold, Stephan and Terwilliger, Jack and Meza, Andres and Zhu, Yixin and Rossano, Federico and Wang, Yizhou},
  journal={International Journal of Computer Vision (IJCV)},
  volume={134},
  number={285},
  year={2026},
  publisher={Springer}
}

@article{ye2024superanimal,
  title={{SuperAnimal} pretrained pose estimation models for behavioral analysis},
  author={Ye, Shaokai and Filippova, Anastasiia and Lauer, Jessy and Schneider, Steffen and Vidal, Maxime and Qiu, Tian and Mathis, Alexander and Mathis, Mackenzie Weygandt},
  journal={Nature Communications},
  volume={15},
  number={5165},
  year={2024},
  publisher={Nature Publishing Group UK London},
  doi = {https://doi.org/10.1038/s41467-024-48792-2},
}

@INPROCEEDINGS{bewley2016sort,
  author={Bewley, Alex and Ge, Zongyuan and Ott, Lionel and Ramos, Fabio and Upcroft, Ben},
  booktitle={2016 IEEE International Conference on Image Processing (ICIP)}, 
  title={Simple online and realtime tracking}, 
  year={2016},
  pages={3464--3468},
  doi={10.1109/ICIP.2016.7533003}}

@misc{blender,
   title = {Blender - a 3D modelling and rendering package},
   author = {{Blender Online Community}},
   organization = {Blender Foundation},
   address = {Blender Institute, Amsterdam},
   year = {2026},
   url = {http://www.blender.org}
 }

@article{eichner2024detailed,
  title={Detailed mapping of the complex fiber structure and white matter pathways of the chimpanzee brain},
  author={Eichner, Cornelius and Paquette, Michael and M{\"u}ller-Axt, Christa and Bock, Christian and Budinger, Eike and Gr{\"a}{\ss}le, Tobias and J{\"a}ger, Carsten and Kirilina, Evgeniya and Lipp, Ilona and Møller, Torsten and Olofsson-Sannö, Karin and Morawski, Markus and Rusch, Henriette and Wenk, Patricia and Weiskopf, Nikolaus and Wittig, Roman M and Crockford, Catherine and Friederici, Angela D and Anwander, Alfred and EBC Consortium and Data Analysis and Writing and Brain Extraction},
  journal={Nature Methods},
  volume={21},
  number={6},
  pages={1122--1130},
  year={2024},
  doi={https://doi.org/10.1038/s41592-024-02270-1}
}

@misc{polyhaven,
  title  = {{Poly Haven} --- Public 3D Asset Library},
  howpublished = {\url{https://polyhaven.com}},
  year   = {2024},
  note   = {Accessed: 2024}
}

@misc{zuffi2019sydog,
      title={{SyDog}: A Synthetic Dog Dataset for Improved {2D} Pose Estimation}, 
      author={Moira Shooter and Charles Malleson and Adrian Hilton},
      year={2021},
      eprint={2108.00249},
}

@article{vogg2025computer,
  title={Computer vision for primate behavior analysis in the wild},
  author={Vogg, Richard and Lueddecke, Timo and Henrich, Jonathan and Dey, Sharmita and Nuske, Matthias and Hassler, Valentin and Murphy, Derek and Fischer, Julia and Ostner, Julia and Schuelke, Oliver and Kappeler, Peter M and Fichtel, Claudia and Gail, Alexander and Treue, Stefan and Scherberger, Hansjörg and Wörgötter, Florentin and Ecker, Alexander S},
  journal={Nature Methods},
  volume={22},
  pages={1154--1166},
  year={2025},
}

@article{kaufman2009can,
  title={Can you believe my eyes? The importance of interobserver reliability statistics in observations of animal behaviour},
  author={Kaufman, Allison B and Rosenthal, Robert},
  journal={Animal Behaviour},
  volume={78},
  number={6},
  pages={1487--1491},
  year={2009},
  doi = {https://doi.org/10.1016/j.anbehav.2009.09.014},
}

@article{mcdougall2012passive,
  title={Is passive observation of habituated animals truly passive?},
  author={McDougall, Petra},
  journal={Journal of Ethology},
  volume={30},
  number={2},
  pages={219--223},
  year={2012},
  publisher={Springer}
}

@article{liu2024enhancing,
title = {Enhancing action recognition from low-quality skeleton data via part-level knowledge distillation},
journal = {Signal Processing},
volume = {221},
pages = {109486},
year = {2024},
issn = {0165-1684},
doi = {https://doi.org/10.1016/j.sigpro.2024.109486},
author = {Cuiwei Liu and Youzhi Jiang and Chong Du and Zhaokui Li},
}

@inproceedings{jiao2025optimizing,
  title={Optimizing Human Pose Estimation Through Focused Human and Joint Regions},
  author={Jiao, Yingying and Wang, Zhigang and Liu, Zhenguang and Fan, Shaojing and Wu, Sifan and Wu, Zheqi and Xu, Zhuoyue},
  booktitle={Proceedings of the AAAI Conference on Artificial Intelligence},
  volume={39},
  number={4},
  pages={4102--4110},
  year={2025}
}

@article{hsu2021b,
  title={{B-SOiD}, an open-source unsupervised algorithm for identification and fast prediction of behaviors},
  author={Hsu, Alexander I and Yttri, Eric A},
  journal={Nature Communications},
  volume={12},
  number={5188},
  year={2021},
  publisher={Nature Publishing Group UK London}
}

@article{luxem2022identifying,
  title={Identifying behavioral structure from deep variational embeddings of animal motion},
  author={Luxem, Kevin and Mocellin, Petra and Fuhrmann, Falko and K{\"u}rsch, Johannes and Miller, Stephanie R and Palop, Jorge J and Remy, Stefan and Bauer, Pavol},
  journal={Communications Biology},
  volume={5},
  number={1267},
  year={2022},
  publisher={Nature Publishing Group UK London}
}

@article{goodwin2024simple,
  title={Simple Behavioral Analysis ({SimBA}) as a platform for explainable machine learning in behavioral neuroscience},
  author={Goodwin, Nastacia L and Choong, Jia J and Hwang, Sophia and Pitts, Kayla and Bloom, Liana and Islam, Aasiya and Zhang, Yizhe Y and Szelenyi, Eric R and Tong, Xiaoyu and Newman, Emily L and Miczek, Klaus and Wright, Hayden R and McLaughlin, Ryan J and Norville, Zane C and Eshel, Neir and Heshmati, Mitra and Nilsson, Simon R O and Golden, Sam A},
  journal={Nature Neuroscience},
  volume={27},
  pages={1411--1424},
  year={2024},
  publisher={Nature Publishing Group US New York}
}

@article{sturman2020deep,
  title={Deep learning-based behavioral analysis reaches human accuracy and is capable of outperforming commercial solutions},
  author={Sturman, Oliver and von Ziegler, Lukas and Schl{\"a}ppi, Christa and Akyol, Furkan and Privitera, Mattia and Slominski, Daria and Grimm, Christina and Thieren, Laetitia and Zerbi, Valerio and Grewe, Benjamin and Bohacek, Johannes},
  journal={Neuropsychopharmacology},
  volume={45},
  number={11},
  pages={1942--1952},
  year={2020},
  publisher={Springer International Publishing Cham}
}

@inproceedings{martini2026bigmaq,
 author = {Martini, Lucas Maximilian and Lappe, Alex and Bogn\'{a}r, Anna and Vogels, Rufin and Giese, Martin},
 booktitle = {International Conference on Learning Representations},
 editor = {C. Vondrick and B. Hariharan and C. Raffel and L. Pinto and D. Yang and A. Faust},
 pages = {117144--117175},
 title = {{BigMaQ}: A Big Macaque Motion and Animation Dataset Bridging Image and {3D} Pose Representations},
 volume = {2026},
 year = {2026}
}

@inproceedings{yang2022apt,
 author = {Yang, Yuxiang and Yang, Junjie and Xu, Yufei and Zhang, Jing and Lan, Long and Tao, Dacheng},
 booktitle = {Advances in Neural Information Processing Systems},
 doi = {10.52202/068431-1258},
 editor = {S. Koyejo and S. Mohamed and A. Agarwal and D. Belgrave and K. Cho and A. Oh},
 pages = {17301--17313},
 publisher = {Curran Associates, Inc.},
 title = {{APT-36K}: A Large-scale Benchmark for Animal Pose Estimation and Tracking},
 volume = {35},
 year = {2022}
}

@article{bala2020automated,
  title={Automated markerless pose estimation in freely moving macaques with {OpenMonkeyStudio}},
  author={Bala, Praneet C and Eisenreich, Benjamin R and Yoo, Seng Bum Michael and Hayden, Benjamin Y and Park, Hyun Soo and Zimmermann, Jan},
  journal={Nature communications},
  volume={11},
  pages={4560},
  year={2020},
  doi = {https://doi.org/10.1038/s41467-020-18441-5}
}

@article{matsumoto2025three,
  title={Three-dimensional markerless motion capture of multiple freely behaving monkeys toward automated characterization of social behavior},
  author={Matsumoto, Jumpei and Kaneko, Takaaki and Kimura, Kei and Blanco Negrete, Salvador and Guo, Jia and Suda-Hashimoto, Naoko and Kaneko, Akihisa and Morimoto, Mayumi and Nishimaru, Hiroshi and Setogawa, Tsuyoshi and Go, Yasuhiro and Shibata, Tomohiro and Nishijo, Hisao and Takada, Masahiko and Inoue, Ken-ichi},
  journal={Science Advances},
  volume={11},
  number={26},
  pages={eadn1355},
  year={2025},
  publisher={American Association for the Advancement of Science},
  doi = {https://doi.org/10.1126/sciadv.adn1355}
}
}

\end{document}